\documentclass[10pt,twocolumn,letterpaper]{article}
\usepackage{multirow}
\usepackage{algpseudocode}
\usepackage{amsmath}
\usepackage{algorithm}
\usepackage{amssymb}
\usepackage{algorithmicx}
\usepackage{booktabs}
\usepackage[margin=1in]{geometry}

\usepackage[pagenumbers]{cvpr} 

\definecolor{cvprblue}{rgb}{0.21,0.49,0.74}
\usepackage[pagebackref,breaklinks,colorlinks,allcolors=cvprblue]{hyperref}

\def\paperID{132} 
\def\confName{CVPR}
\def\confYear{2025}

\title{Enhance Then Search: An Augmentation-Search Strategy with Foundation Models for Cross-Domain Few-Shot Object Detection}

\author{
Jiancheng Pan\textsuperscript{1$\dagger$}, 
Yanxing Liu\textsuperscript{2}, 
Xiao He\textsuperscript{3}, 
Long Peng\textsuperscript{4}, 
Jiahao Li\textsuperscript{1}, 
Yuze Sun\textsuperscript{1}, 
Xiaomeng Huang\textsuperscript{1}\thanks{Xiaomeng Huang is the corresponding author. This work was conducted by the AI4EarthLab team as part of the NTIRE 2025 CD-FSOD Challenge. $\dagger$Jiancheng Pan is the team leader.}\\
\textsuperscript{1}Tsinghua University, 
\textsuperscript{2}University of Chinese Academy of Sciences,\\ 
\textsuperscript{3}Wuhan University,
\textsuperscript{4}University of Science and Technology of China\\
{\tt\small jiancheng.pan.plus@gmail.com}, 
{\tt\small hxm@tsinghua.edu.cn}
}

\begin{document}
\maketitle
\begin{abstract}
Foundation models pretrained on extensive datasets, such as GroundingDINO and LAE-DINO, have performed remarkably in the cross-domain few-shot object detection (CD-FSOD) task. Through rigorous few-shot training, we found that the integration of image-based data augmentation techniques and grid-based sub-domain search strategy significantly enhances the performance of these foundation models. Building upon GroundingDINO, we employed several widely used image augmentation methods and established optimization objectives to effectively navigate the expansive domain space in search of optimal sub-domains. This approach facilitates efficient few-shot object detection and introduces \textbf{an approach to solving the CD-FSOD problem by efficiently searching for the optimal parameter configuration from the foundation model}. Our findings substantially advance the practical deployment of vision-language models in data-scarce environments, offering critical insights into optimizing their cross-domain generalization capabilities without labor-intensive retraining. Code is available at \textcolor{magenta}{https://github.com/jaychempan/ETS}.
\end{abstract}

\section{Introduction}
\begin{figure}[t]
  \centering
  \includegraphics[width=\linewidth]{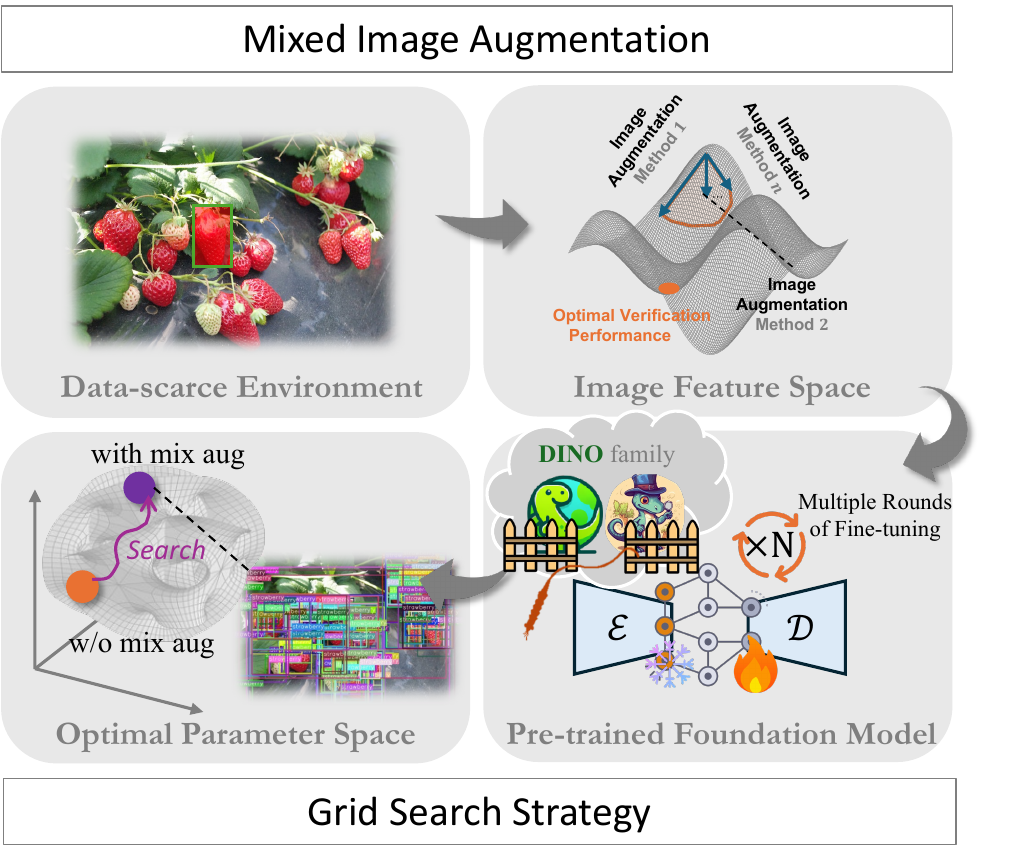}
  \caption{Schematic of proposed augmentation-search strategy for cross-domain few-shot object detection.}
  \label{fig:fig1}
\end{figure}

Recent advancements in foundation models for vision-language tasks, such as GroundingDINO~\cite{liu2024grounding} and LAE-DINO~\cite{pan2024locateearthadvancingopenvocabulary}, have highlighted their exceptional capabilities in zero-shot and few-shot detection across diverse domains. These models, pretrained on large-scale multimodal datasets, exhibit robust generalization abilities for Cross-Domain Few-Shot Object Detection (CD-FSOD) tasks~\cite{fu2024cross,zhuo2022tgdm}. However, the challenge of adapting these models to novel domains with limited annotated samples persists, primarily due to semantic ambiguities that arise in few-shot contexts, such as difficulties in distinguishing visually similar categories and the computational complexities involved in navigating sub-domain spaces. Although various efficient \texttt{pretrain-finetune} methods~\cite{fu2024cross,zhuo2022tgdm,fu2025ntire} have been proposed to address these challenges, the upper-performance limits of foundation models in CD-FSOD scenarios remain primarily unexplored, \textit{\textbf{particularly concerning the synergistic effects of image augmentation and sub-domain adaptation in these data-scarce environments}}.

\begin{figure*}[t]
  \centering
  \includegraphics[width=0.9\linewidth]{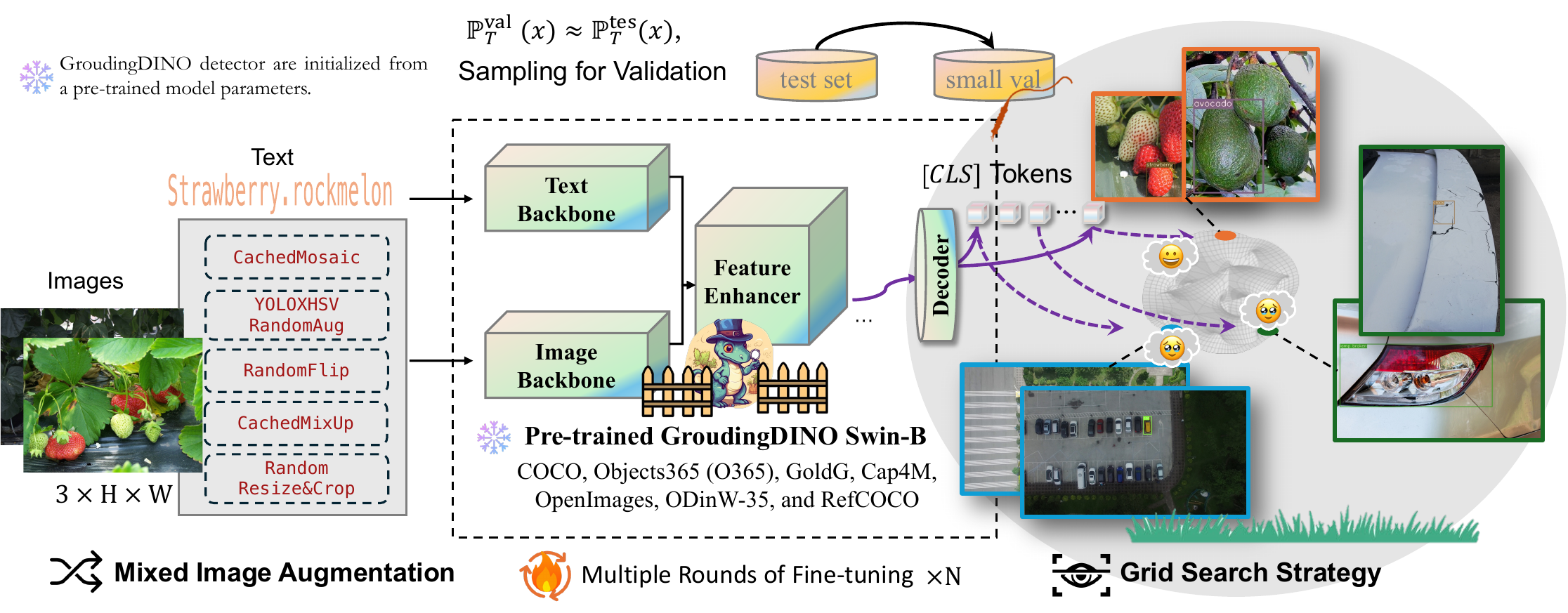}
  \caption{Overall framework of augmentation-search strategy for CD-FSOD, which seamlessly integrates dynamic mixed image augmentation with efficient exploration of domain subspaces.}
  \label{fig:fig2}
\end{figure*}

To address these challenges, we introduce a systematic augmentation-search strategy to unlock the latent potential of vision-language foundation models for the CD-FSOD task. As shown in Figure~\ref{fig:fig1}, our findings indicate that traditional fine-tuning methods often fail to fully exploit the generalization capacity of foundation models, mainly when target-domain data is scarce. For instance, while techniques such as Copy-Paste augmentation may enhance in-domain detection, they can inadvertently introduce instability during cross-domain few-shot adaptation. To mitigate this issue, we have developed a composite augmentation framework that dynamically balances the trade-off between diversity and domain relevance by employing techniques such as \textit{CachedMosaic} and \textit{YOLOXHSVRandomAug}.

Our primary insight lies in conceptualizing CD-FSOD as a joint optimization problem encompassing augmentation policies and domain subspaces. By integrating coarse-grained validation set construction with hyperparameter search mechanisms, we enable the precise adaptation of foundation models, such as GroundingDINO~\cite{liu2024grounding}, with minimal annotation overhead. Experimental results demonstrate that this strategic approach effectively transfers knowledge from diverse pre-training datasets to novel domains, consistently outperforming existing cross-domain object detection baselines in few-shot settings.
The main contributions of this work are threefold:

\begin{itemize}
\item We propose an augmentation-search strategy, \textbf{E}nhance \textbf{T}hen \textbf{S}earch (ETS), which seamlessly integrates mixed image augmentation with efficient exploration of domain subspaces for CD-FSOD.
\item We present an empirical analysis elucidating the intricate interplay between image augmentation diversity, domain shift mitigation, and few-shot detection performance.
\end{itemize}

\section{Related work}
\noindent \textbf{General Object Detection.} With the continuous advancement of deep learning, numerous object detection methods have been proposed, achieving breakthrough progress. These methods can be broadly categorized into two types: two-stage detectors and one-stage detectors.  Among them, a representative two-stage approach is Faster R-CNN~\cite{ren2015faster}, which first generates a set of candidate bounding boxes using a Region Proposal Network (RPN) to identify potential foreground regions. Then, fixed-size feature representations are extracted from each proposal using the RoI pooling module. These object-level features are subsequently classified and refined through bounding box regression to obtain the final detection results. To achieve faster and more accurate detection, many researchers have focused on one-stage detectors. Among these, the YOLO series~\cite{redmon2016you,redmon2018yolov3} is one of the most representative approaches, directly predicting coordinate offsets relative to predefined anchors along with their corresponding classification scores. Recently, with the widespread success of Transformers, numerous transformer-based object detectors~\cite{carion2020end, zhu2020deformable, zhang2022dino} have been proposed, leveraging their powerful representation capabilities to eliminate complex hand-crafted components such as anchors and Non-Maximum Suppression (NMS). Among them, Deformable DETR~\cite{zhu2020deformable} introduces two-dimensional anchor offsets and incorporates a deformable attention module that focuses on specific regions around a reference point. Building on this foundation, DINO~\cite{zhang2022dino} further enhances transformer-based detection by employing a contrastive noise-reduction training strategy, a mixed query selection mechanism, and a look-forward twice scheme. The rapid advancements in natural language processing have also enabled researchers to incorporate rich textual information into object detection. Text-query-based object detection methods~\cite{liu2024grounding, li2022grounded, zareian2021open, radford2021learning, gu2021open} have demonstrated remarkable performance in natural scene understanding. These approaches, trained using existing bounding box annotations, aim to detect arbitrary object categories by leveraging generalized textual queries. OVR-CNN~\cite{zareian2021open} is a pioneering work that pre-trains a visual projector on image-caption pairs to acquire an extensive vocabulary before fine-tuning a detector on a conventional detection dataset. VILD~\cite{gu2021open} enhances open-set object detection by transferring knowledge from the CLIP~\cite{radford2021learning} model into an R-CNN-like detector. GLIP~\cite{li2022grounded} reframes object detection as a grounding task, achieving superior performance on fully supervised detection benchmarks without requiring additional fine-tuning. GroundingDINO~\cite{liu2024grounding} extends GLIP’s training strategy while integrating the robust DINO~\cite{zhang2022dino} detector, establishing new state-of-the-art performance in open-set detection. Furthermore, MQ-Det~\cite{xu2023multi} enhances detection by incorporating both textual and visual queries, effectively leveraging their complementary strengths to improve overall performance.

\begin{figure}[t]
  \centering
  \includegraphics[width=0.95\linewidth]{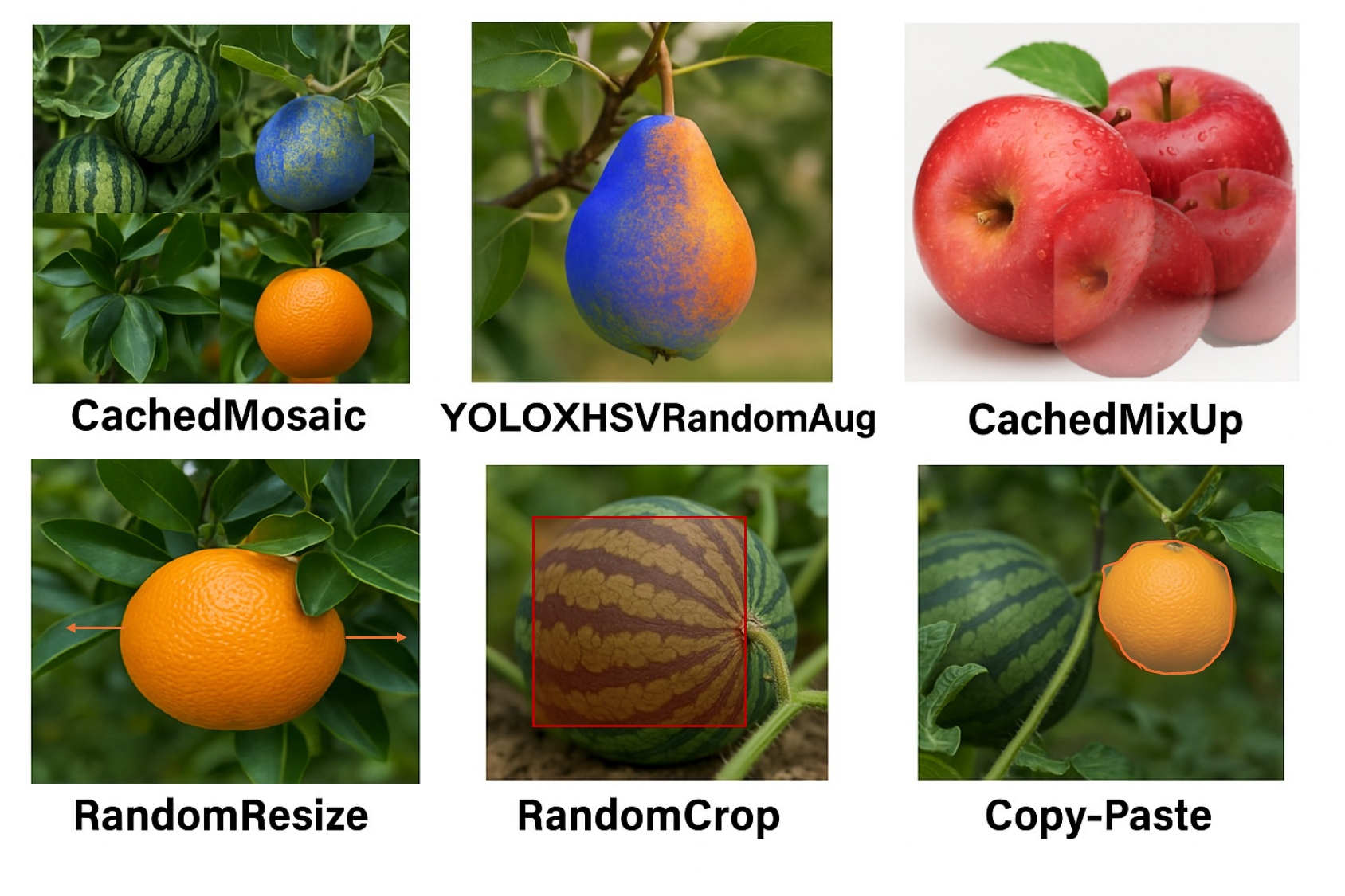}
  \caption{Different image augmentation methods.}
  \label{fig:fig3}
\end{figure}

\noindent \textbf{Few-Shot Object Detection.} Existing FSOD approaches can be categorized into transfer-learning methods~\cite{chen2018lstd,wang2020frustratingly,sun2021fsce,qiao2021defrcn} and meta-learning-based methods~\cite{kang2019few, yan2019meta,xiao2022few,zhang2023detect}. Transfer-learning approaches aim to fine-tune a pre-trained model using limited data for few-shot tasks, with notable works including LSTD~\cite{chen2018lstd}, TFA~\cite{wang2020frustratingly}, FSCE~\cite{sun2021fsce}, and DeFRCN~\cite{qiao2021defrcn}. Meta-learning-based approaches~\cite{kang2019few, yan2019meta,xiao2022few,zhang2023detect} seek to create a class prototype for each category, detecting novel classes by aligning candidate regions with these prototypes. These methods are trained across various tasks to enable rapid adaptation and are optimized over multiple episodes. FSRW~\cite{kang2019few} and Meta-RCNN\cite{yan2019meta} aim to reweight the importance of feature maps by support images to improve the detection performance of few-shot novel classes. Furthermore, FsDetView~\cite{xiao2022few} proposes a novel feature aggregation scheme that uses base class features to enhance detection accuracy on novel classes. DE-ViT\cite{zhang2023detect} introduces a novel approach to open-set object detection by leveraging few-shot learning techniques, utilizing only a small set of support images for each category instead of relying on language-based representations. To further expand the applicability of few-shot object detection, cross-domain few-shot object detection(CD-FSOD) has been proposed. Inspired by cross-domain few-shot learning~\cite{fu2023styleadv,guo2020broader,wang2021cross}, Fu et al.~\cite{fu2024cross} explore the area of CD-FSOD and propose a CD-FSOD benchmark, seeking to develop a robust cross-domain object detector.

\begin{figure}[t]
  \centering
  \includegraphics[width=\linewidth]{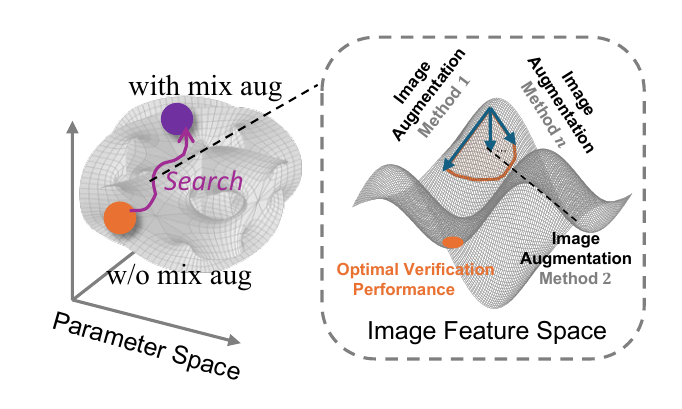}
  \caption{Grid search strategy searches for the optimal mixed image augmentation in the parameter space.}
  \label{fig:fig4}
\end{figure}

\noindent \textbf{Data Augmentation for Object Detection.} Data augmentation, as a widely adopted approach for improving the diversity of training data, has been extensively applied to object detection to enhance performance, particularly in scenarios with limited data. For instance, Kisantal et al.~\cite{kisantal2019augmentation} propose an oversampling strategy for small objects, where images are augmented by copying and pasting small objects multiple times. Building upon this, Ghiasi et al.~\cite{ghiasi2021simple} integrate the copy-paste method with self-supervised learning, further improving detection performance. However, while these augmentation techniques have demonstrated significant effectiveness in limited-data settings, they often fail to generalize well across diverse object distributions and complex real-world scenarios. Moreover, there remains a lack of systematic investigation into their optimal application contexts, interactions with other augmentation strategies, and potential trade-offs between improved diversity and label consistency.

\section{Method}
As shown in Figure~\ref{fig:fig2}, we propose an augmentation-search strategy ETS for the CD-FSOD, which takes open source data and transfers the model to a novel target. This section separately focuses on mixed image augmentation and grid search strategy and describes the overall algorithmic steps.

\subsection{Mixed Image Augmentation}
Different from some data augmentation methods, such as super-resolution\cite{peng2024towards}, which aim to enhance the pixel-level quality of real-world images~\cite{peng2024unveiling,peng2025pixel}, our augmentation strategy focuses on improving the model's robustness under complex conditions. Robustness data augmentation can effectively reduce semantic confusion~\cite{pan2023reducing,pan2023prior,10507076} during few-shot fine-tuning, such as some similar fruit categories in appearance and semantics. Through extensive few-shot fine-tuning, we observe that incorporating image-based data augmentation and optimal domain search strategies can further improve the performance of foundation models, with the upper performance bound remaining unknown. Building upon the open-source GroundingDINO framework, we integrate several commonly used image augmentation techniques and define specific optimization objectives to search for optimal sub-domains within a broad domain space efficiently. This method enables effective few-shot object detection. 

We construct a composite image augmentation pipeline to improve the model's adaptability to various subdomains under limited data scenarios. This pipeline randomly applies a combination of augmentation techniques such as \textit{CachedMosaic}, \textit{YOLOXHSVRandomAug}, \textit{RandomFlip}, \textit{CachedMixUp}, \textit{RandomResize}, and \textit{RandomCrop}. These methods enhance sample diversity, simulate domain shifts, and improve the model’s robustness during fine-tuning. We likewise test more data enhancement methods, including \textit{Copy-Paste}, but we find these methods more unstable during few-shot fine-tuning. As shown in Figure~\ref{fig:fig3}, this pipeline randomly applies a mix of the following techniques:

\begin{itemize}
  \item \textit{CachedMosaic}: Merges four images into a single composite image to expose the model to multiple contexts and object scales within one training sample.
  \item \textit{YOLOXHSVRandomAug}: Introduces photometric variations by adjusting hue, saturation, and value (HSV), simulating different lighting or weather conditions.
  \item \textit{RandomFlip}: Randomly flips the image horizontally and/or vertically to augment training data.
  \item \textit{CachedMixUp}: Blends two images and their labels with a linear combination, encouraging the model to generalize beyond sharp decision boundaries.
  \item \textit{RandomResize}: Dynamically resizes images during training, enabling scale-invariance and improving robustness to varying object sizes.
  \item \textit{RandomCrop}: Randomly crops regions of the image to simulate occlusions and varying viewpoints, helping the model learn from incomplete contexts.
\end{itemize}

\subsection{Grid Search Strategy}
To reduce annotation cost while enabling effective model selection, we construct a validation set $\mathcal{D}_T^{\text{val}}$ by sampling from the annotated target test set $\mathcal{D}_T^{\text{test}} = \{(x_i, y_i)\}_{i=1}^N$, and assigning coarse-grained labels $\tilde{y}_i$ derived from fine annotations $y_i$, i.e.,
\begin{equation}
\mathcal{D}_T^{\text{val}} \sim \mathbb{P}_T(x, \tilde{y}),
\end{equation}
where $\tilde{y}$ retains essential semantics with significantly lower labeling cost. Despite the coarsening of label granularity, we ensure the validation set preserves the input data distribution:
\begin{equation}
\mathbb{P}_T^{\text{val}}(x) \approx \mathbb{P}_T^{\text{test}}(x),
\end{equation}
which guarantees that validation performance provides an unbiased proxy for test-time generalization. This alignment is crucial for reliable hyperparameter optimization under distribution shift.

As shown in Figure~\ref{fig:fig4}, we conduct a grid search over different mixed image augmentation strategies guided by performance on $\mathcal{D}_T^{\text{val}}$. The search identifies optimal parameters that enhance robustness and transferability to the target domain, with optional early stopping to prevent overfitting. Using the selected configuration, we fine-tune the detector and evaluate it on the held-out test set $\mathcal{D}_T^{\text{test}}$. This final inference step offers an unbiased estimate of the model’s domain adaptation performance and confirms the effectiveness of the tuning strategy.

\begin{algorithm}[t]
\caption{Augmentation-Search Strategy based on GroundingDINO}
\label{algorithm:algorithm1}
\begin{algorithmic}[1]

\State \textbf{Model Initialization:} Select foundation model \(\mathcal{M}_{\text{base}} \leftarrow \texttt{GroundingDINO\_SwinB}\), pre-trained on \(\mathcal{D}_{\text{pre}} := \{\texttt{MS-COCO}, \texttt{Objects365}, \ldots\}\).

\State \textbf{Mixed Image Augmentation Pipeline:} Define a mixed image augmentation pipeline with different probability \(\mathcal{A} := \mathcal{A}_{\text{rand}}(\{\texttt{Mosaic}, \texttt{HSV}, \texttt{Flip}, 
 \texttt{MixUp}, \texttt{Resize}, \texttt{Crop}\})\)

\State \textbf{Validation Set Construction:} Build \(\mathcal{D}_{\text{val}} \subset \mathcal{D}_{\text{test}}\) using coarse-grained labels to approximate the target domain.

\State \textbf{Parameter Optimization:} Fine-tune $\mathcal{M}_{\text{base}}$ by searching the parameter space $\Theta$. Identify the optimal configuration by solving:
\[
\theta^* := \arg\max_{\theta \in \Theta} \text{Perf}(\mathcal{M}_\theta, \mathcal{D}_{\text{val}})
\]
where $\text{Perf}(\cdot)$ denotes the performance metric evaluated on the validation set $\mathcal{D}_{\text{val}}$.

\State \textbf{Evaluation:} Load \(\mathcal{M}_{\text{base}}\) with \(\theta^*\) to get \(\mathcal{M}^*\), and evaluate it on \(\mathcal{D}_{\text{test}}\).

\end{algorithmic}
\end{algorithm}

\begin{table*}[t]
\centering
\resizebox{0.75\linewidth}{!}{%
\begin{tabular}{clcccccccc}
\toprule
& \textbf{Method} &  \textbf{Setting} & \textbf{ArTaxOr} & \textbf{Clipart1K} & \textbf{DeepFish} & \textbf{DIOR} & \textbf{NEU-DET} & \textbf{UODD} & \textbf{Avg.}\\
\midrule
\multirow{4}{*}{\rotatebox{90}{1-shot}} 
& Detic ViT-L~\cite{zhou2022detecting} & Close-Source & 3.2 & 15.1 & 9.0 & 4.1 & 3.8 & 4.2 & 6.6 \\
& DE-ViT ViT-L~\cite{zhang2023detect} & Close-Source & 10.5 & 13.0 & 19.3 & 14.7 & 0.6 & 2.4 & 10.1\\
& CD-ViTO ViT-L~\cite{fu2024cross}& Close-Source & 21.0 & 17.7 & 20.3 & 17.8 & 3.6 & 3.1 & 13.9\\
& GroundingDINO* Swin-B~\cite{liu2024grounding} & Open-Source & 22.1 & 56.1 & 37.5 & 11.7 & 10.9 & 19.2 & 26.25 \\
\rowcolor{gray!20}
\rowcolor{gray!20}
& ETS Swin-B (Ours) & Open-Source & 28.1 & 57.5 & 40.7 & 12.7 & 11.7 & 21.2 & 28.65\\
\midrule
\multirow{4}{*}{\rotatebox{90}{5-shot}}
& Detic ViT-L~\cite{zhou2022detecting} & Close-Source & 8.7 & 20.2 & 14.3 & 12.1 & 14.1 & 10.4 & 13.3 \\
& DE-ViT ViT-L~\cite{zhang2023detect} & Close-Source & 38.0 & 38.1 & 21.2 & 23.4 & 7.8 & 5.0 & 22.3\\
& CD-ViTO ViT-L~\cite{fu2024cross} & Close-Source & 47.9 & 41.1 & 22.3 & 26.9 & 11.4 & 6.8 & 26.1\\
& GroundingDINO* Swin-B~\cite{liu2024grounding} & Open-Source & 61.9 & 59.1 & 43.2 & 27.7 & 22.2 & 27.9 & 40.3\\
\rowcolor{gray!20}
\rowcolor{gray!20}
& ETS Swin-B (Ours) & Open-Source & 64.5 & 60.1 & 44.9 & 29.3 & 23.5 & 28.6 & 41.8\\
\midrule
\multirow{4}{*}{\rotatebox{90}{10-shot}}
& Detic ViT-L~\cite{zhou2022detecting} & Close-Source & 12.0 & 22.3 & 17.9 & 15.4 & 16.8 & 14.4 & 16.5 \\
& DE-ViT ViT-L~\cite{zhang2023detect} & Close-Source & 49.2 & 40.8 & 21.3 & 25.6 & 8.8 & 5.4 & 25.2\\
& CD-ViTO ViT-L~\cite{fu2024cross} & Close-Source & 60.5 & 44.3 & 22.3 & 30.8 & 12.8 & 7.0 & 29.6\\
& GroundingDINO* Swin-B ~\cite{liu2024grounding} & Open-Source & 69.8 & 60.8 & 41.5 & 36.4 & 25.4 & 27.4 & 43.6\\
\rowcolor{gray!20}
\rowcolor{gray!20}
& ETS Swin-B (Ours) & Open-Source & 71.2 & 61.5 & 44.1 & 37.5 & 26.1 & 29.8 & 45.0 \\
\bottomrule
\end{tabular}%
}
\caption{The 1/5/10-shot results on six publicly datasets (ArTaxOr, Clipart1K, DeepFish, DIOR, NEU-DET and UODD).}
\label{tab:ad1d2d3}
\end{table*}

\subsection{Augmentation-Search Strategy Procedure}
Algorithm~\ref{algorithm:algorithm1} presents the proposed augmentation-search strategy ETS, which consists of the following steps:

\noindent \textbf{Step 1: Model Initialization.}
We adopt the \textit{Swin-B} version of GroundingDINO as the foundation model, because it is the best within the open-source models. This model has been pre-trained on a diverse set of large-scale datasets, including MS-COCO~\cite{lin2014microsoft}, Objects365~\cite{Shao_2019_ICCV}~\cite{Shao_2019_ICCV}, GoldG~\cite{jenkins2020presentation}, Cap4M~\cite{liu2024grounding}, OpenImages~\cite{kuznetsova2020open}, ODinW-35~\cite{li2022grounded}, and RefCOCO~\cite{kazemzadeh-etal-2014-referitgame}, which collectively provide strong generalization capabilities across multiple vision-language grounding tasks.

\noindent \textbf{Step 2: Mixed Image Augmentation Pipeline.} To improve the model's adaptability to various sub-domains under limited data scenarios, we construct a composite image augmentation pipeline. This pipeline randomly applies a combination of augmentation techniques with probability such as \textit{CachedMosaic}, \textit{YOLOXHSVRandomAug}, \textit{RandomFlip}, \textit{CachedMixUp}, \textit{RandomResize}, and \textit{RandomCrop}. These methods enhance sample diversity, simulate domain shifts, and improve the model’s robustness during fine-tuning. We likewise test more data enhancement methods, including \textit{Copy-Paste}, but we find these methods more unstable during few-shot fine-tuning.

\noindent \textbf{Step 3: Validation Set Construction.} To evaluate adaptation performance, we sample a subset of the annotated test data to serve as a validation set. Instead of full annotations, we apply coarse-grained labeling, which provides enough supervision to guide hyperparameter tuning while reducing annotation cost in the target domain.

\noindent \textbf{Step 4: Parameter Optimization.} We perform hyperparameter search and model selection based on validation performance. This step involves tuning learning rates, augmentation strengths, and other training configurations to identify the optimal setup for domain adaptation.

\noindent \textbf{Step 5: Evaluation.} After identifying the best configuration, we apply the fine-tuned model to the held-out test set to evaluate final performance in the target domain.

\section{Experiment}

\subsection{Dataset and Metric}
The datasets utilized in this paper are categorized into pre-training and fine-tuning datasets that facilitate training.\\
\noindent \textbf{Pre-training Dataset.} Our proposed method builds upon GroundingDINO~\cite{liu2024grounding}, pre-trained on a variety of datasets including MS-COCO~\cite{lin2014microsoft}, Objects365~\cite{Shao_2019_ICCV}~\cite{Shao_2019_ICCV}, GoldG~\cite{jenkins2020presentation}, Cap4M~\cite{liu2024grounding}, OpenImages~\cite{kuznetsova2020open}, ODinW-35~\cite{li2022grounded}, and RefCOCO~\cite{kazemzadeh-etal-2014-referitgame}.\\
\noindent \textbf{Fine-tuning Dataset.} After pre-training, our method establishes a solid foundational capability. To further enhance performance on downstream tasks, we fine-tune the network using publicly available CD-FSOD datasets~\cite{fu2024cross}, including ArTaxOr~\cite{GeirArTaxOr}, Clipart1K~\cite{inoue2018cross}, DIOR~\cite{li2020object}, DeepFish~\cite{saleh2020realistic}, NEU-DET~\cite{song2013noise}, and UODD~\cite{jiang2021underwater}, as well as the unseen datasets DeepFruits~\cite{sa2016deepfruits}, Carpk~\cite{hsieh2017drone}, and CarDD~\cite{wang2023cardd} from the NTIRE 2025 CD-FSOD Challenge~\cite{fu2025ntire}.\\
\noindent \textbf{Evaluation Metric.} To comprehensively validate the effectiveness, we employ two widely used evaluation metrics, including Mean Average Precision (mAP) and Average Precision (AP). Specifically, mAP is a crucial metric for assessing object detection models' performance by quantifying precision and recall across various classes and Intersection over Union (IoU) thresholds.



\subsection{Implementation Details}
All experiments are conducted within the PyTorch framework on eight NVIDIA A100 GPUs. The optimization process incorporates multiple loss functions, including classification and contrastive loss, box L1 loss, and Generalized IoU loss. Following GroundingDINO, Hungarian matching is applied with weight assignments of 2.0 for classification, 5.0 for L1, and 2.0 for GIoU, while the final loss weights are set to 1.0, 5.0, and 2.0, respectively. We employ 900 queries by default and support a maximum text token length of 256. The text encoder is based on BERT. Both the feature enhancer and the cross-modality decoder consist of six layers.  In practice, we tried different learning strategies for different datasets. The learning rate schedule is adjusted using milestone epochs, typically 1, 5, and 9. In the mixed image augmentation methods, the \textit{CachedMosaic}, \textit{RandomFlip}, and \textit{CachedMixUp} strategies are adopted with probabilities of 0.6, 0.5, and 0.3, respectively. In ablation and sampling rate experiments, the fine-tuning experiments are performed on the ArTaxOr dataset without any image augmentation.

\subsection{Detection Results}
We compared the closed-source and the open-source methods with the proposed ETS method. The closed-source method is pre-trained only on the MS-COCO~\cite{lin2014microsoft} dataset, while the open-source method is pre-trained on more datasets for higher performance. All methods are based on fine-tuning and testing to get the final result.

\noindent \textbf{Detection Result on Public Dataset.} The Table~\ref{tab:ad1d2d3} shows the few-shot detection results on ArTaxOr, Clipart1K and DeepFish datasets, comparing proposed ETS method with Detic~\cite{zhou2022detecting}, DE-ViT~\cite{zhang2023detect}, CD-ViTO~\cite{fu2024cross}, and GroundingDINO Swin-B~\cite{liu2024grounding}. To be fairer to GroundingDINO baseline comparisons, GroundingDINO* reports the best mAP result from 4 independent runs, whereas ETS selects the best mAP result from 16 runs or more. GroundingDINO also has a common augmentation method, \textit{RandomResize} and \textit{RandomCrop}. The results of the closed-source approach refer to the original paper by Fu~\cite{fu2024cross}. Compared to the closed-source SOTA method CD-ViTO, our approach achieves average improvements of 14.75, 15.7, and 15.4 percentage points on 1-shot, 5-shot, and 10-shot detection tasks across six datasets, respectively. Compared with the open-source baseline method GroundingDINO, our method yields average gains of 2.4, 1.5, and 1.6 percentage points under the same settings. Overall, open-source methods are better at detecting mAP than closed-source methods. Our ETS consistently achieves better mAP compared to  GroundingDINO on six public datasets.

\begin{table}[t]
\centering
\resizebox{\linewidth}{!}{%
\begin{tabular}{clcccc}
\toprule
& \textbf{Method} &  \textbf{Setting} & \textbf{DeepFruits} & \textbf{Carpk} & \textbf{CarDD} \\
\midrule
\multirow{6}{*}{\rotatebox{90}{1-shot}} 
& CD-ViTO~\cite{fu2024cross} & Close-Source & 28.0 & 6.8 & 10.1 \\
& DFE-ViT~\cite{fu2025ntire} & Close-Source & 32.5 & 18.8 & 18.3 \\
& IFC~\cite{fu2025ntire} & Close-Source & 36.6 & 23.0 & 20.1 \\
& EFT~\cite{fu2025ntire} & Open-Source & 61.3 & 59.2 & 35.1 \\
& PLD~\cite{fu2025ntire} & Open-Source & 63.3 & 61.1 & 32.3 \\
\rowcolor{gray!20}
& ETS (Ours) & Open-Source & 61.2 & 59.2 & 34.2 \\
\midrule
\multirow{6}{*}{\rotatebox{90}{5-shot}} 
& CD-ViTO~\cite{fu2024cross} & Close-Source & 37.4 & 21.3 & 26.5 \\
& DFE-ViT~\cite{fu2025ntire} & Close-Source & 45.2 & 29.4 & 29.1 \\
& IFC~\cite{fu2025ntire} & Close-Source & 47.0 & 29.7 & 29.7 \\
& EFT~\cite{fu2025ntire} & Open-Source & 61.9 & 59.2 & 37.6 \\
& PLD~\cite{fu2025ntire} & Open-Source & 65.4 & 60.4 & 39.2 \\
\rowcolor{gray!20}
& ETS (Ours) & Open-Source & 65.1 & 58.1 & 43.9 \\
\midrule
\multirow{6}{*}{\rotatebox{90}{10-shot}} 
& CD-ViTO~\cite{fu2024cross} & Close-Source & 43.6 & 24.0 & 30.0 \\
& DFE-ViT~\cite{fu2025ntire} & Close-Source & 50.2 & 28.0 & 31.0 \\
& IFC~\cite{fu2025ntire} & Close-Source & 51.0 & 28.0 & 33.0 \\
& EFT~\cite{fu2025ntire} & Open-Source & 64.7 & 59.0 & 40.0 \\
& PLD~\cite{fu2025ntire} & Open-Source & 64.8 & 60.0 & 43.0 \\
\rowcolor{gray!20}
& ETS (Ours) & Open-Source & 65.4 & 59.0 & 47.0 \\
\bottomrule
\end{tabular}%
}
\caption{The 1/5/10-shot results on three unseen datasets (DeepFruits, Carpk, and CarDD).}
\label{tab:d1d2d3}
\end{table}

\noindent \textbf{Detection Result on Unseen Dataset.} Table~\ref{tab:d1d2d3} presents 1-shot, 5-shot, and 10-shot detection results on three unseen datasets, comparing proposed ETS method with CD-ViTO~\cite{fu2024cross}, DFE-ViT~\cite{fu2025ntire}, IFS of the X-Few team~\cite{fu2025ntire},  EFT of the FDUROILab Lenovo team~\cite{fu2025ntire} and PLD of IDCFS team\cite{fu2025ntire} from the NTIRE 2025 CD-FSOD Challenge~\cite{fu2025ntire}. The comparison method is from the challenge, which has been selected for comparison with the same structure, found in the technical report~\cite{fu2025ntire}. Overall, open-source methods are better at detecting mAP than closed-source methods. Our ETS consistently achieves competitive or superior performance across all settings. In the 1-shot scenario, ETS outperforms CD-ViTO significantly and performs on par with PLD, slightly surpassing it on CarDD. For 5-shot, ETS achieves the highest score on CarDD (43.85), while matching PLD on DeepFruits and maintaining comparable results on Carpk. In the 10-shot setting, ETS delivers the best performance on DeepFruits and CarDD, indicating its strong scalability with more shots. These results demonstrate ETS’s robust generalization and adaptability in low-resource cross-domain detection tasks.

\begin{table}[t]
\centering
\resizebox{0.8\linewidth}{!}{%
\begin{tabular}{l c}
\toprule
\textbf{Method} & \textbf{ArTaxOr} (1-shot mAP) \\
\midrule
baseline (w/o aug) & 18.4\\
+ \textit{common aug} & 20.0 (+1.6) \\
+ \textit{mixed aug*} & 22.1 (+3.7) \\
\rowcolor{gray!20}
+ \textit{mixed aug} + \textit{grid search} & 28.1 (+9.7)\\
\bottomrule
\end{tabular}
}
\caption{Ablation studies under the ArTaxOr dataset (1-shot). \textit{mixed aug} is mixed image augmentation, and \textit{grid search} is grid search strategy.}
\end{table}

\subsection{Ablation Study for ETS Method}

To validate the effectiveness of our core idea, we conduct ablation studies on the ArTaxOr dataset, focusing on the impact of mixed image enhancement and grid search within our ETS method. Due to the increased randomness introduced by mixed image augmentation during training, the model's performance exhibits some instability. + \textit{common aug} has a common augmentation method, \textit{RandomResize}
and \textit{RandomCrop}. To better assess the effectiveness of grid search and ensure a fair comparison, + \textit{common aug} and +\textit{mixed aug*} report the best mAP result from 4 independent runs, whereas + \textit{mixed aug} + \textit{grid search} selects the best mAP result from 16 runs. By comparing these results, we observe that a simple augmentation-search strategy can significantly enhance few-shot detection performance, improving up to 9.7 points.

\begin{table}[t]
\centering
\resizebox{0.65\linewidth}{!}{%
\begin{tabular}{c c}
\toprule
\textbf{Sampling Rate} & \textbf{ArTaxOr} (1-shot mAP) \\
\midrule
0.1 & 16.7\\
0.3 & 16.7\\
0.5 & 16.1\\
0.7 & 17.0\\
0.9 & 16.0\\
\rowcolor{gray!20}
Avg. & 16.5\\
\bottomrule
\end{tabular}
}
\caption{Experimental results under the ArTaxOr dataset (1-shot) with sampling rate.}
\end{table}

\subsection{Sampling Rate for Validation}
We conduct the experiments to explore the impact of different sampling rates on validation set construction. To ensure the original distribution of the validation and test sets, we sample in equal proportions by category. The validation set is randomly sampled from the test set, and we evaluate the 1-shot detection performance with milestone epochs set to 1. We observe that varying the sampling ratio has a limited impact on the final optimization results, with performance fluctuations within approximately 16.5. The maximum and minimum mAP values differ from the mean by no more than 0.5. Therefore, sampling a small portion of the test set as the validation set does not affect the final detection results, significantly accelerating the fine-tuning and search process. Although the difficulty of learning on the small sampled validation sets may vary during the sampling process, it does not affect the final optimal performance upon convergence. Even with a small sampling ratio, the optimal results are comparable to those obtained with larger sampling ratios.

\begin{figure}[t]

\centering	{\includegraphics[width=\linewidth]{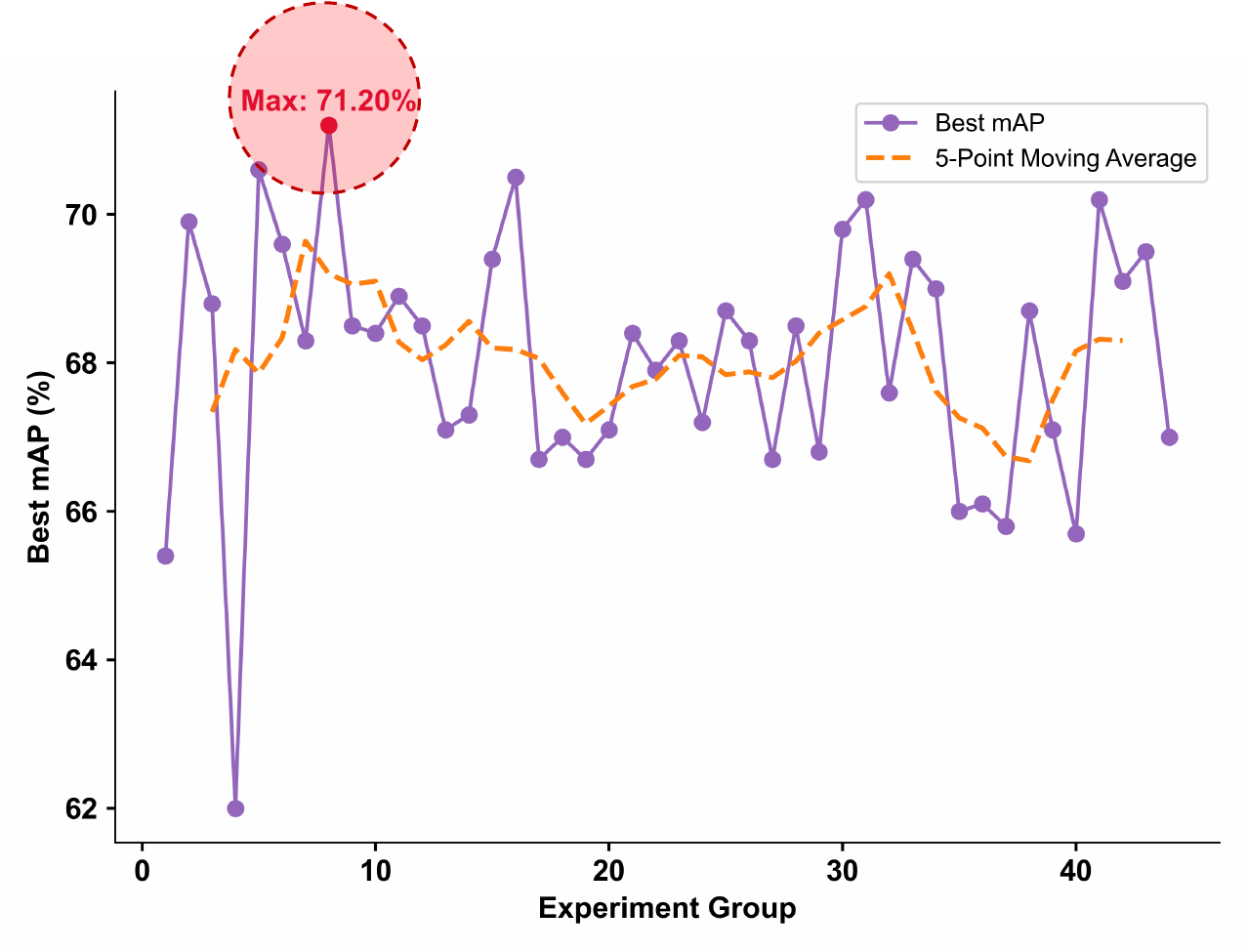}}
\caption{Search strategy experiments for 10-shot detection results on ArTaxOr dataset. \label{fig:exp2} 
}
\end{figure}

\subsection{Search Strategy Feasibility}
As shown in Figure~\ref{fig:exp2}, we validate the necessity of ETS search by performing multiple sets of experiments on the ArTaxOr dataset on 10-shot detection. This set of experimental data includes 44 experiment groups, with the best mAP values ranging from 62.0 to 71.2. The relatively large fluctuations in mAP suggest that finding the best parameter settings within subdomain spaces is an important topic for further CD-FSOD study.

\subsection{Visualisation Results}
\begin{figure}[t]

\centering	{\includegraphics[width=\linewidth]{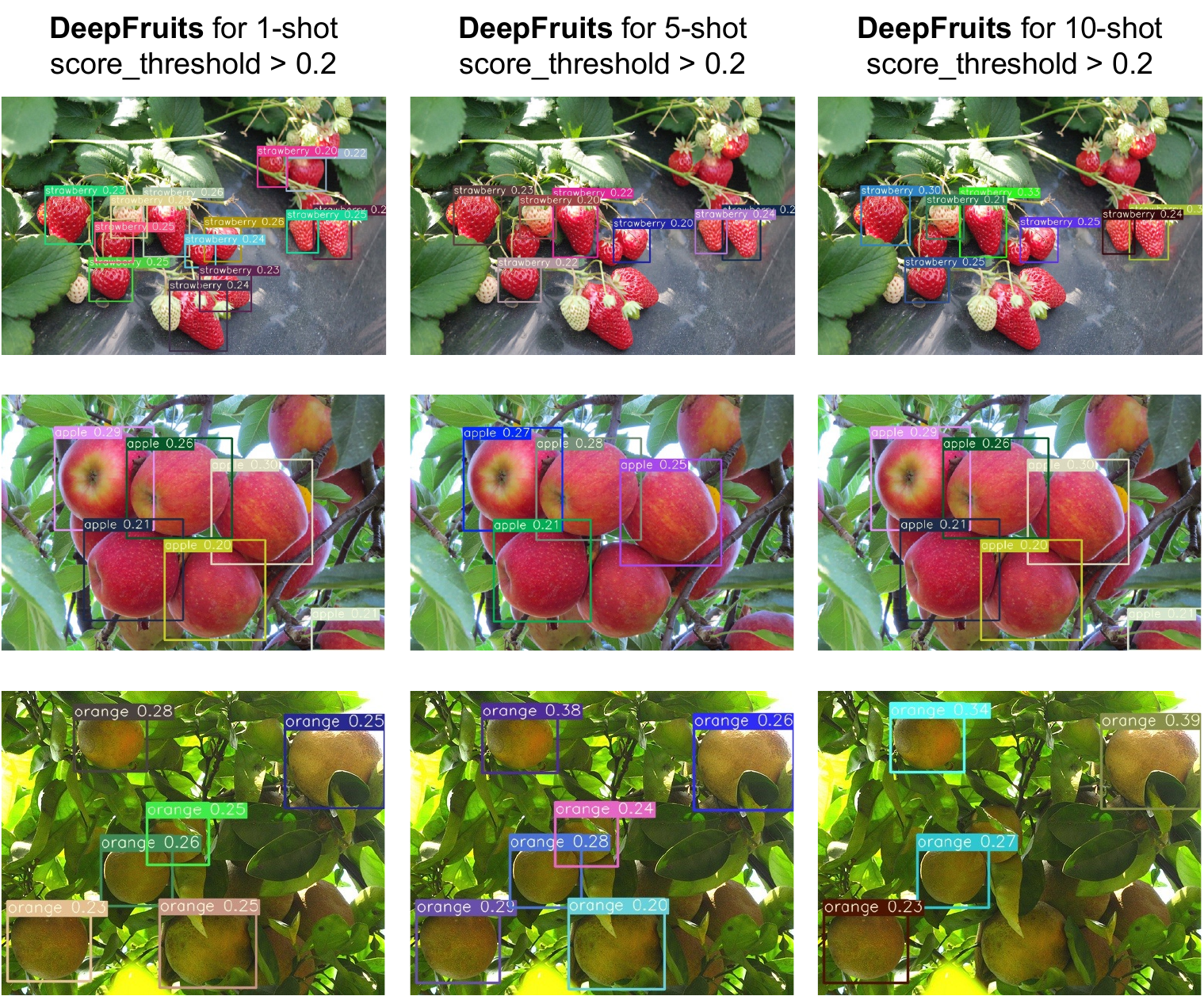}}
\caption{Visualisation for 1-shot detection results on DeepFruits dataset. \label{fig:exp1} 
}
\end{figure}

The Figure~\ref{fig:exp1} shows the visualisation of the 1-shot detection results with different few-shot settings on DeepFruits. It can be observed that compared to the 1-shot setting, the 5-shot and 10-shot settings yield higher accuracy. However, they also discard some reliable detections, focusing more on objects with higher confidence. Therefore, increasing the training samples as much as possible is still a good way to improve the accuracy in real-world object detection tasks.

\section{Conclusion}
This work introduces an augmentation-search strategy, ETS, for CD-FSOD, combining mixed image augmentation with a grid-based sub-domain search strategy. It dynamically balances augmentation diversity and domain relevance using augmentation techniques like \textit{CachedMosaic} and \textit{YOLOXHSVRandomAug}. The strategy achieves consistent performance gains over baselines, enabling better utilization of vision-language models in low-data cross-domain settings. Meanwhile, this work provides a new perspective for addressing the CD-FSOD task by exploring how to efficiently search for the optimal set of parameters from the foundation model. How to further explore the parameter space more effectively in few-shot scenarios to identify optimal parameters remains an open research question.

\section{Acknowledgements}
This work was supported by the National Natural Science Foundation of China (42125503, 42430602).

{
    \small
    \newpage  
    \bibliographystyle{ieeenat_fullname}
    \bibliography{main}
}


\end{document}